\documentclass[conference]{IEEEtran}
\IEEEoverridecommandlockouts
\usepackage{cite}
\usepackage{amsmath,amssymb,amsfonts}
\usepackage{graphicx}
\usepackage{textcomp}
\usepackage{xcolor}
\def\BibTeX{{\rm B\kern-.05em{\sc i\kern-.025em b}\kern-.08em
    T\kern-.1667em\lower.7ex\hbox{E}\kern-.125emX}}

\usepackage{bm}
\usepackage{algorithm}
\usepackage{algpseudocode}
\usepackage{hyperref}
\usepackage{tabularx}
\newcolumntype{x}[1]{!{\centering\arraybackslash\vrule width #1}}
\newcolumntype{L}[1]{>{\raggedright\arraybackslash}p{#1}} % linksbündig mit Breitenangabe

\usepackage{setspace}
\usepackage{adjustbox}
\usepackage{array}

\newcolumntype{R}[2]{%
    >{\adjustbox{angle=#1,lap=\width-(#2)}\bgroup}%
    l%
    <{\egroup}%
}
\newcommand*\rot{\multicolumn{1}{R{70}{1em}}}% no optional argument here, please!

\usepackage{arydshln}

\begin{document}

\title{Impacts of Darwinian Evolution on Pre-trained Deep Neural Networks\\}

% \and
% \and
% \and
% \and
% \and

\author{\IEEEauthorblockN{Guodong Du\IEEEauthorrefmark{2}*, Runhua Jiang\IEEEauthorrefmark{3}*, Senqiao Yang\IEEEauthorrefmark{2}, Haoyang Li\IEEEauthorrefmark{3}, Wei Chen\IEEEauthorrefmark{2}, Keren Li\IEEEauthorrefmark{4}, Sim Kuan Goh\IEEEauthorrefmark{3}, Ho-Kin Tang\IEEEauthorrefmark{2}}
\IEEEauthorblockA{\IEEEauthorrefmark{2}School of Science, Harbin Institute of Technology~(Shenzhen), China.}
\IEEEauthorblockA{\IEEEauthorrefmark{3}School of Electrical Engineering and Artificial Intelligence,
Xiamen University, Malaysia. \\
\IEEEauthorblockA{\IEEEauthorrefmark{4}Center for Quantum Computing, Peng Cheng Laboratory, China.}
*contributed equally to this paper.}
\thanks{This work was supported in part by Shenzhen College Stability Support Plan (GXWD20231128103232001), Department of Science and Technology of Guangdong (2024A1515011540), Shenzhen Start-Up Research Funds~(HA11409065), National Natural Science Foundation of China (12204130), the Ministry of Higher Education Malaysia through the Fundamental Research Grant Scheme (FRGS/1/2023/ICT02/XMU/02/1), and Xiamen University Malaysia through Xiamen University Malaysia Research Fund (XMUMRF/2022-C10/IECE/0039 and XMUMRF/2024-C13/IECE/0049). Corresponding: simkuan.goh@xmu.edu.my, denghaojian@hit.edu.cn}}

\maketitle

\begin{abstract}
Darwinian evolution of the biological brain is documented through multiple lines of evidence, although the modes of evolutionary changes remain unclear. Drawing inspiration from the evolved neural systems (e.g., visual cortex), deep learning models have demonstrated superior performance in visual tasks, among others. While the success of training deep neural networks has been relying on back-propagation (BP) and its variants to learn representations from data, BP does not incorporate the evolutionary processes that govern biological neural systems. This work proposes a neural network optimization framework based on evolutionary theory. Specifically, BP-trained deep neural networks for visual recognition tasks obtained from the ending epochs are considered the primordial ancestors (initial population). Subsequently, the population evolved with differential evolution. Extensive experiments are carried out to examine the relationships between Darwinian evolution and neural network optimization, including the correspondence between datasets, environment, models, and living species. The empirical results show that the proposed framework has positive impacts on the network, with reduced over-fitting and an order of magnitude lower time complexity compared to BP. Moreover, the experiments show that the proposed framework performs well on deep neural networks and big datasets.
\end{abstract}

\begin{IEEEkeywords}
Darwinian evolution, deep learning, neuro-evolution
\end{IEEEkeywords}

\section{Introduction}

\begin{figure*}[h]
\centering
\includegraphics[width=0.9\textwidth]{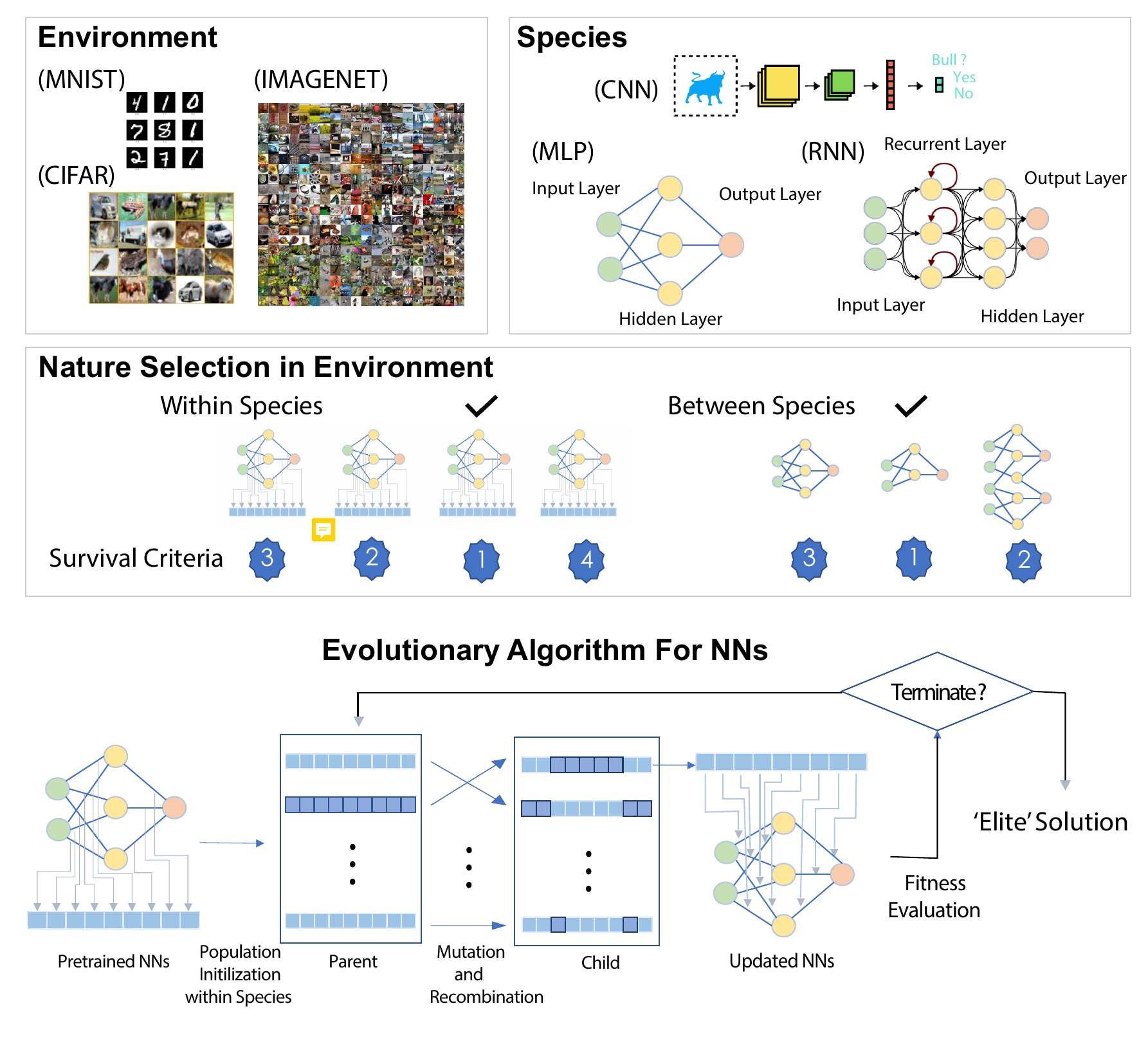}
\caption{\textbf{The conceptual illustration of our proposed Darwinian evolution on DNNs with primordial ancestor.} In analogy to Darwinian evolution, the dataset provides the environment where different types of DNNs strike to survive. The neuro-evolution (natural selection and inheritance) applies to different network architectures, as well as trainable weights in the same architect. Pretrained DNNs are used as the primordial ancestors for EAs to evolve and select the 'elite' solution. The complexity of the EA algorithm is low compared to the backpropagation.} \label{fig1}
\end{figure*}

\begin{figure*}[h]
\centering
\includegraphics[width=0.85\textwidth]{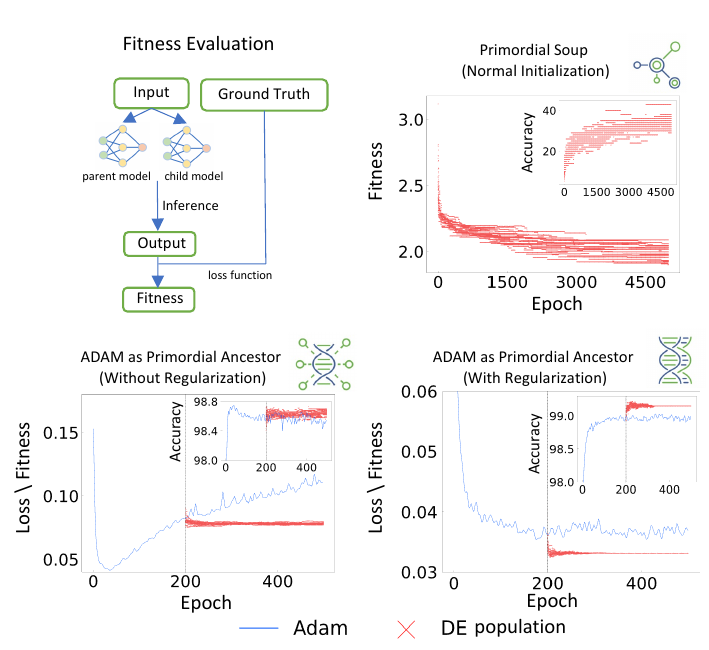}
\caption{\textbf{The starting point of neuro-evolution: primordial ancestor v.s. primordial soup.} The left and middle figures illustrate how the losses and accuracies change over the epoch for DNNs trained by Adam (blue curves) and DE (red markers) using Adam as the primordial ancestor, with and without regularization. It is observed that Adam requires regularization to handle overfitting, while DE does not. The right figure evolves a neural network from the primordial soup using random initialization, which has difficulty in convergence.} \label{fig2}
\end{figure*}

Deep neural networks (DNN)~\cite{LeCun2015-yn,goh2018spatio} have seen remarkable advancements, leading to exceptional performance across a broad spectrum of learning tasks and applications, such as visual tasks~\cite{chen2020simple} and natural language processing~\cite{stiennon2020learning}. Artificial Neural Networks (ANNs) are modeled on the structure and function of the interconnected neurons in the human brain. Training ANNs are equivalent to the search of network weights that optimize a desired loss function, in which intricate network architectures determine the high-dimensional function space. The function landscape is a critical component in determining task performance. For instance, skip connections contribute to the creation of smoother function landscapes in modern network architectures such as RESNET and DENSENET~\cite{Li2018-uj}. Various optimization methods can be employed to search efficiently in the function space. Back-propagation (BP) and its variants (e.g., Adaptive Moment Estimation optimizer (ADAM)~\cite{kingma2014adam}) have established themselves as the most widely used, thanks to their ability to explicitly utilize the gradients of the loss function and enable the training of extremely deep neural networks and Gaussian processes~\cite{9931941,goh2021tunnel}. Nonetheless, BP has several weaknesses~\cite{gong2020evolving}, for example, it requires differentiable loss functions and suffers from the sensitivity to hyperparameters, the problems of vanishing or exploding gradients, slow convergence, and high computational requirements.

An alternative to BP for training DNN is meta-heuristic approaches, for instance, evolutionary algorithms (EA), which implement the theory of evolution~\cite{du2024knowledge, jiang2023enhancing} on computational hardware. EAs have successfully evolved a population of solutions to a plethora of complex optimization problems (e.g., non-convex and NP-hard problems) and real-world problems where traditional methods fail, demonstrating the capability of EAs for general optimization. Utilizing EAs to implement the evolution of DNN architectures and DNN training is termed neuro-evolution. Neuro-evolution bridges DNN optimization and evolutionary theory for algorithmic development, interpretation, and analysis. We illustrate the conceptual framework in Fig.~\ref{fig1}. The neural network architecture and the dataset play the role of the species and environment, respectively. Different architectures specialize in different functions, for example, convolutional neural networks (e.g., ResNet\cite{He2015-ju}, MobileNet\cite{Howard2017-rz}) and recurrent neural networks (e.g., LSTM\cite{low2023lower}, GRU\cite{Chung2014-fm}) capture translation invariances and temporal dependencies underlying the data. The trainable parameters can be interpreted as the genetic traits within the architect framework that affect survival and breeding. On the other side, the complexity of the dataset can be interpreted as the complexity of the environment, scaling from simple MNIST~\cite{Lecun1998-vo} to big data IMAGENET\cite{Deng2009-pv}. Survival fitness can be defined as the loss function of the learning task.

Neuro-evolution has been shown to work well for ANN architecture design and training of reinforcement learning models. Parsimonious neural architectures have been designed through neuro-evolution (e.g., NEAT~\cite{stanley2002evolving}) with enhanced performance. Moreover, neuro-evolution techniques (e.g., evolution strategy~\cite{salimans2017evolution}) have been demonstrated to achieve better results in reinforcement learning tasks compared to deep Q-learning, the policy gradient algorithm A3C~\cite{mnih2016asynchronous} among others. EA-based training of ANNs suffers from very slow (or failure of) convergence, given a large number of model parameters and a complex search space for obtaining the deep representation. Several research works tried to integrate BP with EA-based optimizer by incorporating gradient information but have not been able to show enhanced performance in computer vision tasks the modern testing, like employing deep RESNET in IMAGENET~\cite{o2021evolutionary,sun2019evolving,yang2021gradient}.

% The EA-based training of ANNs suffers from very slow (or failure of) convergence, given a large number of model parameters and a complex search space for obtaining the deep representation. Several research works tried integrating BP with an EA-based optimizer by incorporating gradient information. Still, they failed to show consistently enhanced performance in computer vision tasks that have been revolutionized by deep learning, with neural networks of different depths and data with different complexities~\cite{o2021evolutionary,sun2019evolving,yang2021gradient}. Here, we re-investigate the problem and study how gradient-based and EA-based methods can be used together to train ANNs better. Specifically, we consider a neural network as an existing species and investigate the impacts of Darwinian evolution on it.

% The EA-based training of ANNs suffers from very slow (or failure of) convergence, given a large number of model parameters and a complex search space for obtaining the deep representation. Several research works tried to integrate BP with EA-based optimizer by incorporating gradient information but have not been able to show enhanced performance in computer vision tasks the modern testing, like employing deep RESNET in IMAGENET ~\cite{o2021evolutionary,sun2019evolving,yang2021gradient}.

Here, we re-investigate the problem and explore the feasibility of integrating BP-based and EA-based methods. Our approach conceptualizes the neural network as a species and examines the effects of Darwinian evolution on its performance. Using pre-trained DNNs as parent models, we discovered that EA optimization could further enhance the accuracy of DNNs in image classification tasks. To thoroughly examine the efficacy of our proposed framework, we conducted extensive experiments considering various factors such as the selection of parent models, dataset complexity, and DNN architecture depth, demonstrating a correlation between datasets, environment, models, and living species and superior performance in deep neural networks and large datasets. Given the low computational complexity of EAs, our framework presents a practical and effective solution for improving the performance of ANNs following traditional gradient-based training.

\section{Background}
The proposed work differs from other EA-based methods that train ANNs from random initialization, which imitate the primordial soup (i.e., prior to the formation of the first species, primordial ancestor, also known as the last universal common ancestor (LUCA)). Primordial soup theory~\cite{taylor2005stirring}, Miller-Urey experiment~\cite{miller1953production}, and others~\cite{kasting1993earth} studied and simulated the conditions of the early Earth for the first life, which arose from non-living matters, give rise to other species through evolution. The initialization that implements primordial soup can be the reason for the slow convergence due to the unmet conditions for the evolutionary starting point. Inspired by Hinton's work of pretraining the neural network using restricted Boltzmann machines to avoid the vanishing of gradient, we use BP-based optimizer (i.e., ADAM) for the pretraining of DNN and EA-based methods to fine-tune DNN's weights. We consider the pretrained DNNs, obtained from the ending epochs, as the primordial ancestor (i.e., the first species), the starting point of evolution. Subsequently, Darwinian evolution is implemented using differential evolution. BP-based algorithms allow a single neural network to accumulate knowledge and learn representation from data, while neuro-evolution emulates the evolution of a population of neural networks. The same strategy applies to other pretrained models, facilitating the evolution of fitter breeds (i.e., enhanced pre-trained models). Several computer vision datasets and DNN architects are used to validate the concepts. This work focuses on evolving a population of trainable parameters in the pre-defined architecture (i.e., single organism) rather than evolving a population of different architectures~\cite{real2017large,lu2020multiobjective}.

\section{Proposed Method}
Consider a task to learn a neural network with dataset $\mathcal{D}$, where each sample $\bm{x}\in \mathcal{D}$ is with multi-features and a label $y$. A fundamental question is:

\emph{How can we obtain a neural network that faithfully depicts the training set and accurately predicts the testing set? (or can we say "How can we learn a good neural network?" here.)}

Assumed that training and testing data follow an identical distribution. 
   Let $\bm{\theta}$ denote the parameters of the neural network model.
In the standard learning process, optimal parameters can be found by minimizing loss $\mathcal{L}$, which is formulated as 
\begin{eqnarray}
    \bm{\theta}=\arg \min_{\bm{\theta}} \mathcal{L}(\bm{\theta}; \bm{x}, y), \label{eq:obj}
\end{eqnarray}
Back-propagation and gradient-based methods are almost ubiquitously for such minimization with elaborately designed regularization $\Omega(\bm{\theta})$. 
% However, the proliferation of local minima 
Based on that, we provide an alternative method by evolutionary algorithm, which is specified as two stages in Fig.~\ref{fig1} and Algorithm~\ref{alg:EaLearning} shows the typical steps.

The first stage is initialization with back-propagation and gradient-based methods, which is shown at the bottom of Fig.~\ref{fig1}. 
This is an iterative algorithm that starts with a random input and fitness evaluation.
For each epoch, gradient-based quantity is calculated and $\bm{\theta}$ can thus be updated until the termination condition is satisfied. 

This produces a set of candidate solutions,
\begin{eqnarray}
    \Theta=\{\bm{\theta}_i: i=1,...,m\},
\end{eqnarray}
where $m$ is called population size, and $\bm{\theta}_i=(\theta_{i,1},\theta_{i,2},...,\theta_{i,d})$, $d$ is the size of solution space.

The second stage is to boost with an evolutionary algorithm, which generates new solutions from the current candidate set.
The algorithm is shown at the bottom of Fig.~\ref{fig1}, where differential evolution is taken as an example.
This is also an iterative algorithm that starts with $\Theta$.
For each generation(epoch), one step is to generate “mutant” vectors, which form $\Theta^{\star}$,  
\begin{eqnarray}
   \bm{\theta}^{\star}_{i}=\bm{\theta}_{j}+F\times (\bm{\theta}_{k}-\bm{\theta}_{l}), 
\end{eqnarray}
where $i=1,...,m$ and $j$, $k$, $l$ are random integers less than $m$, different from $i$ and other. $F$ is a tunable scalar.  
Then, recombination is performed, which includes crossover and selection. Crossover is taken place with a pre-set threshold $Cr$. For $j=1,...d$,
\begin{eqnarray}
   \theta^{\star}_{i,j}  = \begin{cases}
        \theta^{\star}_{i,j} & \mbox{if} \quad rand(0,1)\leq Cr,\\
        \theta_{i,j} & \mbox{otherwise}.
     \end{cases}
\end{eqnarray}
For selection, the choice of $\bm{\theta}_i$ in the next generation is made between $ \bm{\theta}^{\star}_{i}$ and $\bm{\theta}_{i,j}$ 
\begin{eqnarray}
    \bm{\theta}_i = \begin{cases}
        \bm{\theta}^{\star}_{i} & \mathcal{L}(\bm{\theta}^{\star g}_{i}) <  \mathcal{L}(\bm{\theta}^{g}_i) \\
        \bm{\theta}_{i} & \mbox{otherwise}
     \end{cases}
\end{eqnarray}
where the lowest loss function is targeted by direct one-to-one comparison.

\begin{algorithm}[!ht]
    \begin{algorithmic}[1]
       \Procedure{EaLearning}{$\mathcal{D}$,t,m}
          \State $\bm{\theta} \gets \textsc{RandomInput}$     \Comment{random input}
          \State $\Delta f\gets 1$
          \State $f\gets \textsc{Loss}(\bm{\theta},\mathcal{D}, t)$ \Comment{evaluation}
          \While {$f \geq \delta_1 \land \Delta f \geq \Delta_1$} 
             \State $\hat{g} \gets \textsc{BatchBackProp}(\mathcal{D},t)$ \Comment{gradient calculation}
             \State $\bm{\theta}\gets \textsc{Update}(\bm{\theta},\hat{g})$ \Comment{updating parameters}
             \State $f'\gets \textsc{Loss}(\bm{\theta}, \mathcal{D}, t)$ 
             \State $\Delta f\gets f-f'$
             \State $f\gets f'$  
             \State $\Theta \gets \textsc{Gets}(\bm{\theta},m)$     \Comment{store recent m $\bm{\theta}$}
          \EndWhile
          \While {$f \geq \delta_2 \land \Delta f \geq \Delta_2$}
             \State $\Theta^{\star} \gets \textsc{Mutate}(\Theta)$ \Comment{mutation}
             \State $\Theta \gets \textsc{Recombination}(\Theta,\Theta^{\star})$ \Comment{crossover and selection}
             \For {$\bm{\theta}_i\in \Theta$}            
             \State $f'_i\gets \textsc{Loss}(\bm{\theta}_i, \mathcal{D}, t)$ 
             \State $\Delta f_i\gets f-f'_i$ 
             \EndFor
             \State $(f,\Delta f,\bm{\theta}) \gets \textsc{Min}(\Delta f_i , f_i, \Theta)$ \Comment{choose the optimal $\bm{\theta}$}
          \EndWhile
          \State\Return $\bm{\theta}$
       \EndProcedure
    \end{algorithmic}
    \caption{\textsc{EaLearning}: algorithm for learning optimal $\theta$ of neural network. Here $\mathcal{D}$ is the training dataset, $t$ is the type of networks, and $m$ is the population size. $\delta_1$, $\delta_2$, $\Delta_1$ and $\Delta_2$ are pre-set threshold.}
    \label{alg:EaLearning}
 \end{algorithm}
Remarkably, if a $m$-layers network is employed, $\sum_{i=1}^m l_i $ parameters are to be optimized, where $l_i$ represents number of $i$-th layer.
Time complexity is thus $\mathcal{O}(n_{g}\cdot\prod_{i=1}^{m-1}l_il_{i+1})$ and $\mathcal{O}(n_{e}\cdot\sum_{i=1}^{m}l_i)$  where $n_g$ and $n_e$ are training samples for gradient-based back-propagation and evolutionary algorithms, respectively. 

Step by step, we analyze the algorithmic complexity for both gradient-based back propagation and evolutionary algorithm to train a layered neural network. 

We emphasize that our analysis is on training a $m$-layered neural network, where $\sum_{i=1}^m l_i $ parameters are to be optimized. The procedure to produce updated parameters is focused. 

For feed-forward pass direction, each layer has experienced such a process
\begin{eqnarray}
   Z_{i+1} \leftarrow M_{i+1,i}\cdot Z_i, \quad Z_{i+1} \leftarrow f(Z_{i+1}),
\end{eqnarray}
where $f(*)$ is the activation function and $M_{i+1,i}$ contains the weights going from layer $i$ to $i+1$.
Thus, time complexity is the same as feed-forward case, which in total demands $\mathcal{O}({n_{g}\sum_{i=1}^{m-1}l_il_{i+1}})$ basic operations and $\mathcal{O}({n_{g}\sum_{i=1}^{m-1}l_{i+1}})$ queries to the inverse activation function.

For back-propagation direction, each layer has experienced such a process
\begin{eqnarray}
    E_{i} \leftarrow f'(Z_{i}-O_{i}), D_{i,i-1}\leftarrow E_{i} \cdot Z_{i-1}, \\ M_{i,i-1} \leftarrow  M_{i,i-1}- D_{i,i-1}
\end{eqnarray}
where $E_{i-1}$ and $D_{i,i-1}$ are the error terms and adjust matrix. Remarkably, different algorithms are supposed to be employed here and we consider the typical case.  
Thus, time complexity is $\mathcal{O}({n_{g}l_il_{i+1}})$ operations, and $\mathcal{O}({n_{g}l_{i}})$ queries to the inverse activation function.
In total, feed-forward pass algorithm demands $\mathcal{O}({n_{g}\sum_{i=1}^{m-1}l_il_{i+1}})$ basic operations and $\mathcal{O}({n_{g}\sum_{i=1}^{m-1}l_{i+1}})$ queries to the activation function.

For the differential evolution algorithm, Eq. (3), (4), and (5) demand $\mathcal{O}(n_e \sum_{i=1}^{m-1}l_i)$.

\section{Experiment}

\begin{figure*}[h]
\centering
\includegraphics[width=0.85\textwidth]{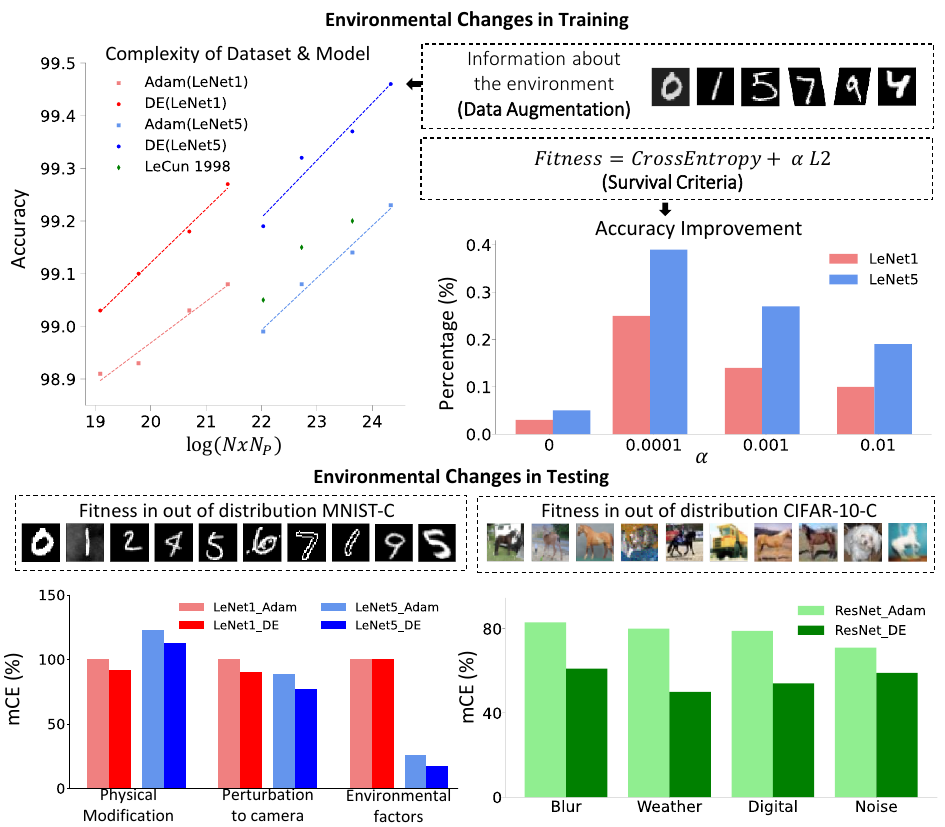}
\caption{\textbf{The impact of environmental changes on the neural network during training and testing.} The upper left figure shows the relationships between the accuracy and the complexity of the dataset (MNIST with data augmentation) and model (LeNet). As complexity increases, the accuracy increases for both Adam and DE. DE is observed to achieve higher accuracy compared to ADAM. The upper right figure illustrates the influence of regularization, which has a positive impact on improving accuracy. The two figures at the bottom show the performance of ADAM and DE on the corrupted and out-of-distribution MNIST-C and CIFAR-10-C, using LeNet1, LeNet5, and ResNet. DE is found to have a lower mean corruption error than Adam.} \label{fig3}
\end{figure*}

\begin{figure*}[h]
\centering
\includegraphics[width=0.95\textwidth]{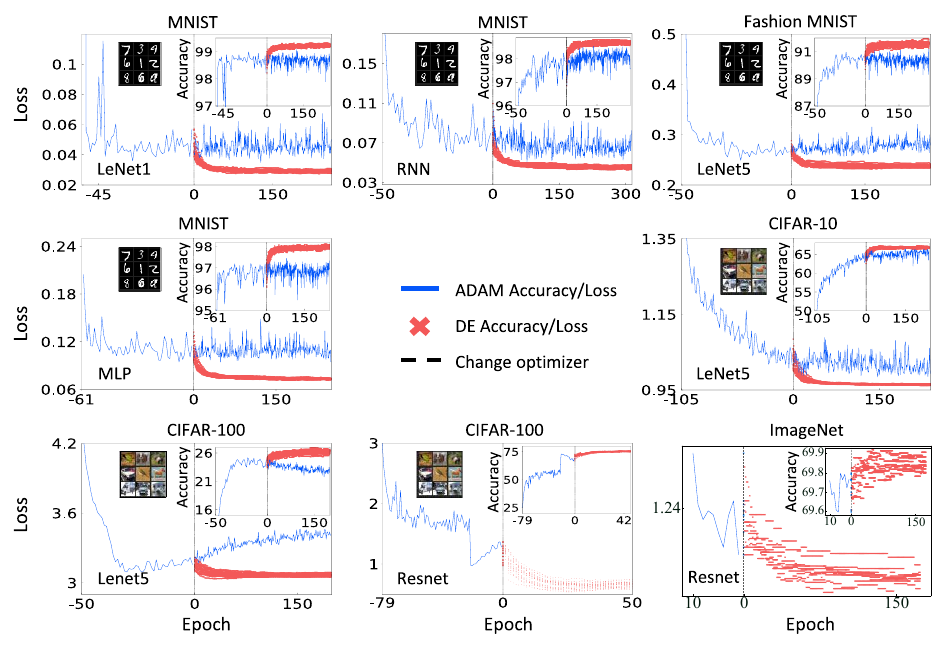}
\caption{\textbf{Generalization of DE on different datasets and deep learning models.} Datasets include MNIST, Fashion MNIST\cite{Xiao2017-hq}, CIFAR-100, and ImageNet, while models include: LeNet1, LeNet5, MLP, RNN, ResNet18. The blue lines indicate Adam as the optimizer, and the red points indicate DE as the optimizer.} \label{fig4}
\end{figure*}

%So we are interested to check whether other algorithms can perform well, and this also inspire neuroevolution, mainly on designs DNN’s architectures, and relatively limited works on using evolutionary algorithms (EA) to optimize.

%So far how to evolve the architect has been a fruitful research direction, but we face a lot of constraint in subtle genetic mutations within each architect. In evolution sense, the accumulation of genetic changes is the source of the evolution of organism, the thorough hyperparameter search guiding the architect search by checking which kernel is more active or involved in the evolution.

% Elites cultivation -  ADAM
% Inheritance -  DE
% Adapat environment L2/no L2 or dataset(same solution different dataset)
% Natural selection in DE <-> parameter mutation rate
% Applicability to all NN and dataset
% Time complexity
% E.A. v.s. B.P. insight

% =>Inheritance evolution - regularized /not regularized , regularized solve convergence problem.
% -> anti-over-fitting (change to without L2)

Here, BP-trained deep neural networks using ADAM~\cite{kingma2014adam}, obtained from the ending epochs, are considered the primordial ancestors in the Darwinian evolution of a population of networks for visual recognition tasks. The neural networks are optimized using cross-entropy with $L2$ regularization as the loss function. During Darwinian evolution, natural selection is implemented using cross-entropy as a fitness function that determines the relative survival and reproductive success of a species.

Extensive experiments are carried out on models such as LeNet1, LeNet5, and a larger ResNet18 trained on MNIST, FashionMNIST, Cifar-10, Cifar-100, and ImageNet. Additional verification of model robustness is assessed on MNIST-C and CIFAR-10-C datasets.

\subsection{Hyper-parameters}
Grid search is employed to identify the optimal hyperparameters, and the specified search range is outlined in Table~\ref{tab:A2}.

\begin{table}[!htbp]
  \begin{center}
    \caption{Hyper-parameter search space}
    \label{tab:A2}
    \begin{tabular}{lc}
      \hline
      Hyperparameters & Search Space                     \\ \hline

      mutate       & {[}0.01, 0.1, 1, 2{]} \\ \hline
      recombination       & {[}0, 0.05, 0.5, 1{]}              \\ \hline
      learning rate   & {[}1e-2, 2e-2{]}           \\ \hline
    \end{tabular}
  \end{center}
  % \label{hyper}
\end{table}

\section{Result and Discussion}
Empirically, neuro-evolution that begins with the primordial ancestor is found to be superior to the primordial soup, as shown in Fig.~\ref{fig2}. Hence, it shows the importance of the formation of a species before evolution takes place.

Comparing the primordial ancestor trained with and without regularization, it is also observed that the quality of the primordial ancestor has a huge impact on DNNs, demonstrating the effect on descent in the same lineage. In the same lineage, offspring receive the genetic traits from parents through inheritance. Offspring inherit major traits with parents, however, with slight variation and modification. In Fig.~\ref{fig2}, the lineage trained using BP with regularization performs better. The top right of Fig.~\ref{fig3} shows the different degrees of improvement that can be achieved on top of the BP-based approach under different degrees of $L2$ regularization. It is observed that the best results are achieved with a regularization of 0.0001 compared to no regularization at all, and all regularization parameters lead to increased model accuracy. The use of regularization reduces the impact of over-fitting in BP-based optimizers. During the Darwinian evolution, it is observed that the trait of preventing over-fitting is inherited without an explicit $L2$ regularization in the fitness function.

% This further allows for a more significant increase in the model's accuracy at the DE stage.

% Inheritance is the process that allows offspring to receive the genetic traits from parents, through mutation, recombination, etc. Offspring inherits major traits with parents, however, with their slight variation and modification. Similarly, in the training of ANN, the offspring population inherits the traits of the parent solution. In Fig.~\ref{fig2}, we show that the quality of the initialized population has a huge impact on using the evolutionary algorithm, demonstrating the effect on descent in the same lineage. 

% At the same time, gradient-based optimization algorithms can quickly collect and acquire large amounts of knowledge, while evolutionary algorithms can improve the robustness and generalization performance of the model. By combining gradient-based optimization with evolutionary algorithms, the models can perform better and be more resilient in complex environments. 

In Fig.\ref{fig3}, LeNet1 and LeNet5 architectures are used to simulate two different species in the environment (dataset) with different complexities, which provide different amounts of training data generated using data augmentation. The former represents the primitive life form, namely a simpler model with fewer parameter counts, while the latter represents the higher form of living species. The Darwinian evolution is observed to have positive impacts on the ADAM-trained neural network. Moreover, datasets and models with higher complexity are observed to achieve better performance. Both LeNet1 and LeNet5 evolved using the proposed Darwinian evolution framework are found to be better than the LeNet5 trained using BP by LeCun et al. in 1998~\cite{Lecun1998-vo}. Numerical results are provided in Table~\ref{tab:numresult}

% on both LeNet1 and LeNet5 is found to be 

% After the Darwinian evolution on the neural network models, results on the top left corner of the Fig.\ref{fig3} shows that  

% are the same for these two species(without data augmentation), our algorithm achieve good performance on both LeNet1 and LeNet5. Both results were better than what LeCun et al. achieved in 1998.

% We simulated the environmental change by changing the complexity of the dataset, which was equivalent to increasing the initial individuals within a specie. It increases the chance of discovering more excellent parents in each specie. And our result reflect this conclusion: as the complexity of data increased, the accuracy rate also increased.

% The survival condition of individual models throughout the environment is the fitness function. And the well-adapted models have higher classification accuracy. The top right of Fig.\ref{fig3} shows the different degrees of improvement that can be achieved by our method compared to the gradient-based approach under different degrees of L2 regularization. It can be observed that the best results are achieved with a regularization of 0.0001 compared to no regularization at all. Including regularization in the survival environment ensures the model has a better generalization ability (in other words, a better gene). This further allows for a more significant increase in the model's accuracy at the DE stage.

To assess the out-of-distribution robustness of the neural network trained using the proposed framework, two datasets with common corruptions are used, namely MNIST-C and CIDAR-10-C. The types of corruption are summarized at the bottom of Fig.~\ref{fig3} as well as detailed in Tables III to V. From Tables III and IV, LeNet5 trained by DE is found to be less susceptible to corruption, demonstrating the positive impacts of DE on BP-trained shallow networks. Similar positive impacts on network robustness under corruptions are observed for deeper neural networks, ResNet and MobileNet, in Table V.

% . Firstly, MNIST-C contains 15 types of image corruptions. In several experiments and visualizations, this corruption captures failure patterns of models not previously found in the literature. By preserving the semantic content of the underlying images, this corruption increases the error rate of convolutional neural networks by a factor of 10 compared to clean data. 

% Another one is Cifar10-C. As a robustness benchmark in computer vision, Cifar10-C contains 15 common visual corruptions with five severity levels. The results show that the mCE(mean corruption error) of the trained model by DE is lower than that of the trained model by Adam, so it is clear that the trained model by DE is more stable.

To further explore the generalization capabilities of the proposed framework, deeper models and big datasets are used. As shown in the bottom rightmost sub-figure in Fig.~\ref{fig4}, it is found that using ResNet18 on cifar100\cite{Krizhevsky2009-st} and ImageNet also leads to decreases in validation loss, increases in validation accuracy, and improvements in overall model performance. As shown in Table~\ref{tab:numresult}, the positive impacts of the proposed optimization framework on classification accuracies of test sets are shown. It further demonstrates that our approach is not only applicable to small models but also achieves good improvements on big datasets and deep models. 

% Due to the limitation of computational resources, we only tried three sets of parameters each, but we can already see the improvement of ResNet18 on ImageNet.

According to the results, the proposed framework shows enhanced classification performance when compared with BP-based methods; no overfitting problem is observed in EA training, demonstrating advantageous effects similar to regularization; low time complexity compared to the back-propagation, making it highly practical to be incorporated into the recent framework of DNN training.

% \begin{table}[h]
% \setlength{\tabcolsep}{4pt}
% \renewcommand{\arraystretch}{1}
% \tiny\centering
% \caption{\textbf{Test accuracy.} This table summarizes the performance of Adam and DE on different deep learning models and datasets.} 
% \begin{tabular}{l|ccccccc}
% % & \multicolumn{7}{c|}{\textbf{Error}} & \multicolumn{9}{c}{\textbf{mCE}} \\
% \\
%   & {ResNet18 ImageNet} & {LeNet1  MNIST} & {LeNet5  MNIST}& {LeNet5  Fashion MNIST}& {LeNet5  Cifar10} & {LeNet5 Cifar100} & {ResNet18 Cifar100} \\ \hline
% Adam & 66.12\% & 98.78\%  & 98.95\% & 89.27\% & 60.16\% & 24.47\%   & 70.83\%  \\
% DE & {\color[HTML]{003399} \textbf{66.87\%}} & {\color[HTML]{003399} \textbf{99.03\%}} & {\color[HTML]{003399} \textbf{99.20\%}}& {\color[HTML]{003399} \textbf{89.43\%}}& {\color[HTML]{003399} \textbf{61.36\%}} & {\color[HTML]{003399} \textbf{26.73\%}}& {\color[HTML]{003399} \textbf{72.89\%}}\\

% \end{tabular}
% \end{table}

\begin{table}[h]
\setlength{\tabcolsep}{4pt}
\renewcommand{\arraystretch}{1}
\centering

\caption{\textbf{Test accuracy.} This table summarizes the performance of Adam and DE on different deep learning models and datasets.} \label{tab:numresult}
\begin{tabular}{l|cccc}
& Adam & DE & parameters & datasets\\ \hline
{LeNet1  MNIST} & 98.78 & {\color[HTML]{003399} 99.03} & 3,246   & 60M \\
{LeNet5  MNIST} & 98.95 &  {\color[HTML]{003399} 99.20} &  62,006  & 60M  \\
{LeNet5  Fashion MNIST} & 89.27 & {\color[HTML]{003399} 89.43}&  62,006  & 80M  \\
{LeNet5  Cifar10} & 60.16 &  {\color[HTML]{003399} 61.36}& 62,006 &  180M \\
{LeNet5  Cifar100} & 24.47 & {\color[HTML]{003399} 26.73} & 62,006 &  180M \\
{ResNet18 Cifar100} & 70.83 &{\color[HTML]{003399} 72.89} & 11,220,132  & 180M  \\
{ResNet18 ImageNet} & 66.12 & {\color[HTML]{003399} 66.87} & 11,220,132  & 150GB \\

\end{tabular}
\end{table}

Other than the inheritance of elites' anti-overfitting property and the improved solution by natural selection, the proposed framework also has low time complexity compared to BP-based optimizers.
If a $m$-layers network is employed, $\sum_{i=1}^m l_i $ parameters are to be optimized, where $l_i$ represents number of $i$-th layer.
Time complexity is thus $\mathcal{O}(n_{g}\cdot\sum_{i=1}^{m-1}l_il_{i+1})$ and $\mathcal{O}(n_{e}\cdot\sum_{i=1}^{m}l_i)$  where $n_g$ and $n_e$ are training samples for gradient-based BP and evolutionary algorithms, respectively. 
The scheme is also readily available for many existing models, as we can use the existing pretrained model as the initialized population. Though the scheme's improvement in performance is of a modest scale, this represents a milestone in using EAs to train DNNs. We expect more advanced schemes to be discovered in this direction.

\subsection{Models' robustness against corruption}
To verify the robustness and generalization of our approach, we trained the model using MNIST data and Cifar10 data and tested it on MNIST-C and Cifar10-C datasets, respectively. Table~\ref{tab1} shows LeNet1 and LeNet5's performance on MNIST-C. Table~\ref{tab2} shows LeNet1 and LeNet5's performance on Cifar10-C. Table~\ref{tab3} shows ResNet performance on Cifar10-C. Generally, it is observed that DE enhanced DNN robustness for almost all models.

\begin{table}[h]
\setlength{\tabcolsep}{4pt}
\renewcommand{\arraystretch}{1}
\tiny\centering
\caption{\textbf{Performance on MNIST-C.} 
The error and mean corruption error (mCE) of ADAM and DE for LeNet models trained on MNIST. Models that perform the best for each type of corruption are colored blue. Overall, LeNet5 trained by DE is found to be less susceptible to corruption.} 
\label{tab1}
\begin{tabular}{l|rrrrr|rrrr}
& \multicolumn{4}{c}{\textbf{Error}} & & \multicolumn{4}{c}{\textbf{mCE}}\\
 &\rot{LeNet1\_Adam} & \rot{LeNet1\_DE} & \rot{LeNet5\_Adam} & \rot{LeNet5\_DE} &  & \rot{LeNet1\_Adam} & \rot{LeNet1\_DE} & \rot{LeNet5\_Adam} & \rot{LeNet5\_DE} \\ \hline
Shot Noise & 2.7\% & 2.4\% & 3.0\% & {\color[HTML]{003399} \textbf{2.1\%}} &  & 100\% & 92\% & 114\% & {\color[HTML]{003399} \textbf{79\%}} \\
Impulse Noise & 13.6\% & 14.2\% & 11.3\% & {\color[HTML]{003399} \textbf{9.6\%}} &  & 100\% & 104\% & 83\% & {\color[HTML]{003399} \textbf{70\%}} \\
Glass Blur & 13.1\% & 12.2\% & 8.9\% & {\color[HTML]{003399} \textbf{6.3\%}} &  & 100\% & 93\% & 68\% & {\color[HTML]{003399} \textbf{48\%}} \\
Motion Blur & 20.1\% & 15.6\% & 10.8\% & {\color[HTML]{003399} \textbf{8.9\%}} &  & 100\% & 78\% & 54\% & {\color[HTML]{003399} \textbf{44\%}} \\
Shear & 3.5\% & 3.1\% & 3.1\% & {\color[HTML]{003399} \textbf{2.5\%}} &  & 100\% & 89\% & 88\% & {\color[HTML]{003399} \textbf{74\%}} \\
Scale & 11.2\% & {\color[HTML]{003399} \textbf{7.9\%}} & 12.4\% & 8.1\% &  & 100\% & {\color[HTML]{003399} \textbf{71\%}} & 111\% & 73\% \\
Rotate & 9.6\% & 8.7\% & 9.5\% & {\color[HTML]{003399} \textbf{7.9\%}} &  & 100\% & 90\% & 99\% & {\color[HTML]{003399} \textbf{83\%}} \\
Brightness & 84.9\% & 84.8\% & 8.2\% & {\color[HTML]{003399} \textbf{4.2\%}} &  & 100\% & 100\% & 10\% & {\color[HTML]{003399} \textbf{5\%}} \\
Translate & 44.9\% & 43.0\% & 43.3\% & {\color[HTML]{003399} \textbf{42.0\%}} &  & 100\% & 96\% & 96\% & {\color[HTML]{003399} \textbf{94\%}} \\
Stripe & 32.2\% & 31.8\% & 15.1\% & {\color[HTML]{003399} \textbf{8.2\%}} &  & 100\% & 99\% & 47\% & {\color[HTML]{003399} \textbf{26\%}} \\
Fog & 81.9\% & 82.4\% & 22.0\% & {\color[HTML]{003399} \textbf{18.2\%}} &  & 100\% & 101\% & 27\% & {\color[HTML]{003399} \textbf{22\%}} \\
Spatter & 5.2\% & 4.9\% & 3.1\% & {\color[HTML]{003399} \textbf{2.6\%}} &  & 100\% & 95\% & 59\% & {\color[HTML]{003399} \textbf{49\%}} \\
Dotted Line & 4.8\% & 4.6\% & 4.0\% & {\color[HTML]{003399} \textbf{3.6\%}} &  & 100\% & 96\% & 84\% & {\color[HTML]{003399} \textbf{76\%}} \\
Zigzag & 16.0\% & 15.4\% & 13.0\% & {\color[HTML]{003399} \textbf{11.3\%}} &  & 100\% & 96\% & 81\% & {\color[HTML]{003399} \textbf{71\%}} \\
Canny Edges & 22.9\% & {\color[HTML]{003399} \textbf{20.1\%}} & 40.2\% & 37.9\% &  & 100\% & {\color[HTML]{003399} \textbf{88\%}} & 176\% & 165\% \\
Average & 24.4\% & 23.4\% & 13.9\% & {\color[HTML]{003399} \textbf{11.6\%}} &  & 100\% & 92\% & 80\% & {\color[HTML]{003399} \textbf{65\%}}
\end{tabular}
\end{table}

\begin{table}[h]
\setlength{\tabcolsep}{4pt}
\renewcommand{\arraystretch}{1}
\tiny\centering
\caption{\textbf{Performance on CIFAR10-C.} The error and mean corruption error (mCE) of ADAM and DE for LeNet models trained on CIFAR10. Models that perform the best for each type of corruption is colored blue. In all corruptions, LeNet5 trained by DE is found to be less susceptible to corruption.} \label{tab2}
\begin{tabular}{l|rrrrr|rrrr}
& \multicolumn{4}{c}{\textbf{Error}} & & \multicolumn{4}{c}{\textbf{mCE}}\\
 &\rot{LeNet1\_Adam} & \rot{LeNet1\_DE} & \rot{LeNet5\_Adam} & \rot{LeNet5\_DE} &  & \rot{LeNet1\_Adam} & \rot{LeNet1\_DE} & \rot{LeNet5\_Adam} & \rot{LeNet5\_DE} \\ \hline
Gaussian & 52.7\% & 52.3\% & 44.4\% & {\color[HTML]{003399} \textbf{43.7\%}} &  & 100\% & 99\% & 84\% & {\color[HTML]{003399} \textbf{83\%}} \\
Shot & 51.2\% & 51.2\% & 42.5\% & {\color[HTML]{003399} \textbf{42.3\%}} &  & 100\% & 100\% & 83\% & {\color[HTML]{003399} \textbf{83\%}} \\
Impulse & 56.2\% & 56.1\% & 48.3\% & {\color[HTML]{003399} \textbf{47.4\%}} &  & 100\% & 100\% & 86\% & {\color[HTML]{003399} \textbf{84\%}} \\
Defocus & 51.3\% & 49.9\% & 44.2\% & {\color[HTML]{003399} \textbf{43.4\%}} &  & 100\% & 97\% & 86\% & {\color[HTML]{003399} \textbf{85\%}} \\
Glass & 53.3\% & 52.1\% & 48.0\% & {\color[HTML]{003399} \textbf{46.9\%}} &  & 100\% & 98\% & 90\% & {\color[HTML]{003399} \textbf{88\%}} \\
Motion & 53.2\% & 51.9\% & 49.0\% & {\color[HTML]{003399} \textbf{47.4\%}} &  & 100\% & 97\% & 92\% & {\color[HTML]{003399} \textbf{89\%}} \\
Zoom & 53.5\% & 52.1\% & 48.5\% & {\color[HTML]{003399} \textbf{46.9\%}} &  & 100\% & 97\% & 91\% & {\color[HTML]{003399} \textbf{88\%}} \\
Snow & 52.1\% & 52.1\% & 46.4\% & {\color[HTML]{003399} \textbf{46.0\%}} &  & 100\% & 100\% & 89\% & {\color[HTML]{003399} \textbf{88\%}} \\
Frost & 54.9\% & 55.9\% & 52.4\% & {\color[HTML]{003399} \textbf{50.2\%}} &  & 100\% & 102\% & 96\% & {\color[HTML]{003399} \textbf{92\%}} \\
Fog & 58.8\% & 57.6\% & 56.7\% & {\color[HTML]{003399} \textbf{54.7\%}} &  & 100\% & 98\% & 96\% & {\color[HTML]{003399} \textbf{93\%}} \\
Brightness & 50.8\% & 49.8\% & 44.4\% & {\color[HTML]{003399} \textbf{43.5\%}} &  & 100\% & 98\% & 87\% & {\color[HTML]{003399} \textbf{86\%}} \\
Contrast & 67.2\% & 65.5\% & 65.3\% & {\color[HTML]{003399} \textbf{64.0\%}} &  & 100\% & 97\% & 97\% & {\color[HTML]{003399} \textbf{95\%}} \\
Elastic & 52.7\% & 51.7\% & 45.8\% & {\color[HTML]{003399} \textbf{45.0\%}} &  & 100\% & 98\% & 87\% & {\color[HTML]{003399} \textbf{85\%}} \\
Pixel & 49.7\% & 49.1\% & 41.7\% & {\color[HTML]{003399} \textbf{41.1\%}} &  & 100\% & 99\% & 84\% & {\color[HTML]{003399} \textbf{83\%}} \\
JPEG & 49.7\% & 48.7\% & 42.0\% & {\color[HTML]{003399} \textbf{41.1\%}} &  & 100\% & 98\% & 84\% & {\color[HTML]{003399} \textbf{83\%}} \\
Average & 53.8\% & 53.1\% & 48.0\% & {\color[HTML]{003399} \textbf{46.9\%}} &  & 100\% & 99\% & 89\% & {\color[HTML]{003399} \textbf{87\%}}
\end{tabular}

\end{table}

\begin{table}[h]
\setlength{\tabcolsep}{4pt}
\renewcommand{\arraystretch}{1}
\tiny\centering
\caption{\textbf{Performance of deeper models on CIFAR10-C.} The error and mean corruption error (mCE) of ADAM and DE for deeper models trained on CIFAR10. Models that perform the best for each type of corruption are colored blue. Overall, deeper models trained by DE are found to be less susceptible to corruption. ResNet and MobileNet are observed to be comparable, where each manages certain corruptions better.} 
\label{tab3}
\begin{tabular}{l|rrrrr|rrrr}
% & \multicolumn{7}{c|}{\textbf{Error}} & \multicolumn{9}{c}{\textbf{mCE}} \\
& \multicolumn{4}{c}{\textbf{Error}} & & \multicolumn{4}{c}{\textbf{mCE}}\\
  &\rot {ResNet\_Adam} & \rot{ResNet\_DE} & \rot{MobileNet\_Adam} & \rot{MobileNet\_DE} &  & \rot{ResNet\_Adam} & \rot{ResNet\_DE} & \rot{MobileNet\_Adam} & \rot{MobileNet\_DE} \\ \hline
Gaussian & 37.5\% & {\color[HTML]{003399} \textbf{30.9\%}} & 68.6\% & 68.1\% &  & 71\% & {\color[HTML]{003399} \textbf{59\%}} & 130\% & 129\% \\
Shot & 36.0\% & {\color[HTML]{003399} \textbf{28.4\%}} & 56.2\% & 55.6\% &  & 70\% & {\color[HTML]{003399} \textbf{55\%}} & 110\% & 109\% \\
Impulse & 39.9\% & {\color[HTML]{003399} \textbf{35.8\%}} & 49.2\% & 49.1\% &  & 71\% & {\color[HTML]{003399} \textbf{64\%}} & 88\% & 87\% \\
Defocus & 40.3\% & 26.9\% & 21.7\% & {\color[HTML]{003399} \textbf{21.2\%}} &  & 79\% & 52\% & 42\% & {\color[HTML]{003399} \textbf{41\%}} \\
Glass & 42.4\% & {\color[HTML]{003399} \textbf{37.8\%}} & 49.6\% & 48.9\% &  & 80\% & {\color[HTML]{003399} \textbf{71\%}} & 93\% & 92\% \\
Motion & 49.3\% & 33.8\% & 31.1\% & {\color[HTML]{003399} \textbf{30.3\%}} &  & 93\% & 64\% & 58\% & {\color[HTML]{003399} \textbf{57\%}} \\
Zoom & 44.0\% & 30.7\% & 28.9\% & {\color[HTML]{003399} \textbf{28.3\%}} &  & 82\% & 57\% & 54\% & {\color[HTML]{003399} \textbf{53\%}} \\
Snow & 41.1\% & 29.6\% & 22.2\% & {\color[HTML]{003399} \textbf{21.8\%}} &  & 79\% & 57\% & 43\% & {\color[HTML]{003399} \textbf{42\%}} \\
Frost & 45.4\% & 27.6\% & 28.5\% & {\color[HTML]{003399} \textbf{27.9\%}} &  & 83\% & {\color[HTML]{003399} \textbf{50\%}} & 52\% & 51\% \\
Fog & 47.4\% & 30.3\% & 17.8\% & {\color[HTML]{003399} \textbf{17.4\%}} &  & 81\% & 51\% & 30\% & {\color[HTML]{003399} \textbf{30\%}} \\
Brightness & 39.5\% & 21.4\% & 9.3\% & {\color[HTML]{003399} \textbf{9.2\%}} &  & 78\% & 42\% & 18\% & {\color[HTML]{003399} \textbf{18\%}} \\
Contrast & 61.9\% & 43.7\% & 33.1\% & {\color[HTML]{003399} \textbf{32.6\%}} &  & 92\% & 65\% & 49\% & {\color[HTML]{003399} \textbf{49\%}} \\
Elastic & 41.6\% & 28.0\% & 20.5\% & {\color[HTML]{003399} \textbf{20.0\%}} &  & 79\% & 53\% & 39\% & {\color[HTML]{003399} \textbf{38\%}} \\
Pixel & 36.7\% & {\color[HTML]{003399} \textbf{24.7\%}} & 26.4\% & 26.9\% &  & 74\% & {\color[HTML]{003399} \textbf{50\%}} & 53\% & 54\% \\
JPEG & 36.0\% & {\color[HTML]{003399} \textbf{23.0\%}} & 24.0\% & 24.0\% &  & 72\% & {\color[HTML]{003399} \textbf{46\%}} & 48\% & 48\% \\
Average & 42.6\% & {\color[HTML]{003399} \textbf{30.2\%}} & 32.5\% & 32.1\% &  & 79\% & {\color[HTML]{003399} \textbf{56\%}} & 61\% & 60\%
\end{tabular}
\end{table}

\subsection{Impact of mutate and recombination}
In nature, where natural selection already rules out the evolution strategy that is too extreme, as they are extinct already.
So in DE, evolutionary processes such as inheritance with modification (i.e., mutation and recombination) and natural selection of fitter individuals are implemented.
% As shown in Fig.~\ref{fig5} and Fig.~\ref{fig6}, 
Empirically, the evolution of DNN solutions only occurs when the mutation and recombination are not too big nor too small, requiring the inheritance of advantageous traits from the parents while maintaining diversity in the population.

\section{Conclusion}

All in all, the proposed framework has shown three important traits compared to BP-based optimizer: the generally enhanced performance, the low time complexity, and the most important one -- stability. (ADAM\cite{kingma2014adam}: will over-fit the training data when regularization is not used, DE: does not require explicit regularization to prevent over-fitting and has good performance against noise.) Stability matters most in living to maintain survival and reproduction from generation to generation. In contrast, the extreme strategy can 
cause extinction, which might explain why the brain does not explicitly compute the gradient for BP. According to our results and findings, the strategy that uses a pre-trained neural network by BP to accumulate knowledge and learn representation from data then evolves the network using the proposed framework, provides a practical and robust approach to enhanced network performance. Future research can explore adaptive DE for deeper neural networks trained on datasets with larger scales. 
Github URL: \url{https://github.com/Yangsenqiao/Impacts-of-Darwinian-Evolution-on-Deep-Neural-Networks}.

\bibliographystyle{IEEEtran}
\bibliography{sn-bibliography}

\begin{thebibliography}{10}
\providecommand{\url}[1]{#1}
\csname url@samestyle\endcsname
\providecommand{\newblock}{\relax}
\providecommand{\bibinfo}[2]{#2}
\providecommand{\BIBentrySTDinterwordspacing}{\spaceskip=0pt\relax}
\providecommand{\BIBentryALTinterwordstretchfactor}{4}
\providecommand{\BIBentryALTinterwordspacing}{\spaceskip=\fontdimen2\font plus
\BIBentryALTinterwordstretchfactor\fontdimen3\font minus \fontdimen4\font\relax}
\providecommand{\BIBforeignlanguage}[2]{{%
\expandafter\ifx\csname l@#1\endcsname\relax
\typeout{** WARNING: IEEEtran.bst: No hyphenation pattern has been}%
\typeout{** loaded for the language `#1'. Using the pattern for}%
\typeout{** the default language instead.}%
\else
\language=\csname l@#1\endcsname
\fi
#2}}
\providecommand{\BIBdecl}{\relax}
\BIBdecl

\bibitem{LeCun2015-yn}
Y.~LeCun, Y.~Bengio, and G.~Hinton, ``Deep learning,'' \emph{Nature}, 2015.

\bibitem{goh2018spatio}
S.~K. Goh, H.~A. Abbass, K.~C. Tan, A.~Al-Mamun, N.~Thakor, A.~Bezerianos, and J.~Li, ``Spatio--spectral representation learning for electroencephalographic gait-pattern classification,'' \emph{IEEE Transactions on Neural Systems and Rehabilitation Engineering}, vol.~26, no.~9, pp. 1858--1867, 2018.

\bibitem{chen2020simple}
T.~Chen, S.~Kornblith, M.~Norouzi, and G.~Hinton, ``A simple framework for contrastive learning of visual representations,'' in \emph{International conference on machine learning}.\hskip 1em plus 0.5em minus 0.4em\relax PMLR, 2020, pp. 1597--1607.

\bibitem{stiennon2020learning}
N.~Stiennon, L.~Ouyang, J.~Wu, D.~Ziegler, R.~Lowe, C.~Voss, A.~Radford, D.~Amodei, and P.~F. Christiano, ``Learning to summarize with human feedback,'' \emph{Advances in Neural Information Processing Systems}, vol.~33, pp. 3008--3021, 2020.

\bibitem{Li2018-uj}
H.~Li, Z.~Xu, G.~Taylor, C.~Studer, and T.~Goldstein, ``Visualizing the loss landscape of neural nets,'' \emph{Adv. Neural Inf. Process. Syst.}, vol.~31, 2018.

\bibitem{kingma2014adam}
D.~P. Kingma and J.~Ba, ``Adam: A method for stochastic optimization,'' \emph{arXiv preprint arXiv:1412.6980}, 2014.

\bibitem{9931941}
S.~K. Goh, N.~P. Singh, Z.~J. Lim, and S.~Alam, ``Interpretable tracking and detection of unstable approaches using tunnel gaussian process,'' \emph{IEEE Transactions on Aerospace and Electronic Systems}, vol.~59, no.~3, pp. 2658--2671, 2023.

\bibitem{goh2021tunnel}
S.~K. Goh, Z.~Jun~Lim, S.~Alam, and N.~Pratap~Singh, ``Tunnel gaussian process model for learning interpretable flight’s landing parameters,'' \emph{Journal of Guidance, Control, and Dynamics}, vol.~44, no.~12, pp. 2263--2275, 2021.

\bibitem{gong2020evolving}
M.~Gong, J.~Liu, A.~K. Qin, K.~Zhao, and K.~C. Tan, ``Evolving deep neural networks via cooperative coevolution with backpropagation,'' \emph{IEEE Transactions on Neural Networks and Learning Systems}, vol.~32, no.~1, pp. 420--434, 2020.

\bibitem{du2024knowledge}
G.~Du, J.~Li, H.~Liu, R.~Jiang, S.~Yu, Y.~Guo, S.~K. Goh, and H.-K. Tang, ``Knowledge fusion by evolving weights of language models,'' \emph{arXiv preprint arXiv:2406.12208}, 2024.

\bibitem{jiang2023enhancing}
R.~Jiang, S.~Yang, H.~Li, H.~Wang, H.-K. Tang, and S.~K. Goh, ``Enhancing deep neural network corruption robustness using evolutionary algorithm,'' in \emph{2023 IEEE International Conference on Cybernetics and Intelligent Systems (CIS) and IEEE Conference on Robotics, Automation and Mechatronics (RAM)}.\hskip 1em plus 0.5em minus 0.4em\relax IEEE, 2023, pp. 55--60.

\bibitem{He2015-ju}
K.~He, X.~Zhang, S.~Ren, and J.~Sun, ``Deep residual learning for image recognition,'' pp. 770--778, Dec. 2015.

\bibitem{Howard2017-rz}
A.~G. Howard, M.~Zhu, B.~Chen, D.~Kalenichenko, W.~Wang, T.~Weyand, M.~Andreetto, and H.~Adam, ``{MobileNets}: Efficient convolutional neural networks for mobile vision applications,'' Apr. 2017.

\bibitem{low2023lower}
W.~S. Low, K.~Y. Goh, S.~K. Goh, C.~H. Yeow, K.~W. Lai, S.~L. Goh, J.~H. Chuah, and C.~K. Chan, ``Lower extremity kinematics walking speed classification using long short-term memory neural frameworks,'' \emph{Multimedia Tools and Applications}, vol.~82, no.~7, pp. 9745--9760, 2023.

\bibitem{Chung2014-fm}
J.~Chung, C.~Gulcehre, K.~Cho, and Y.~Bengio, ``Empirical evaluation of gated recurrent neural networks on sequence modeling,'' Dec. 2014.

\bibitem{Lecun1998-vo}
Y.~Lecun, L.~Bottou, Y.~Bengio, and P.~Haffner, ``Gradient-based learning applied to document recognition,'' \emph{Proc. IEEE}, vol.~86, no.~11, pp. 2278--2324, Nov. 1998.

\bibitem{Deng2009-pv}
J.~Deng, W.~Dong, R.~Socher, L.~J. Li, K.~Li, and {others}, ``Imagenet: A large-scale hierarchical image database,'' \emph{2009 IEEE conference}, 2009.

\bibitem{stanley2002evolving}
K.~O. Stanley and R.~Miikkulainen, ``Evolving neural networks through augmenting topologies,'' \emph{Evolutionary computation}, vol.~10, no.~2, pp. 99--127, 2002.

\bibitem{salimans2017evolution}
T.~Salimans, J.~Ho, X.~Chen, S.~Sidor, and I.~Sutskever, ``Evolution strategies as a scalable alternative to reinforcement learning,'' \emph{arXiv preprint arXiv:1703.03864}, 2017.

\bibitem{mnih2016asynchronous}
V.~Mnih, A.~P. Badia, M.~Mirza, A.~Graves, T.~Lillicrap, T.~Harley, D.~Silver, and K.~Kavukcuoglu, ``Asynchronous methods for deep reinforcement learning,'' in \emph{International conference on machine learning}.\hskip 1em plus 0.5em minus 0.4em\relax PMLR, 2016, pp. 1928--1937.

\bibitem{o2021evolutionary}
D.~O’Neill, B.~Xue, and M.~Zhang, ``Evolutionary neural architecture search for high-dimensional skip-connection structures on densenet style networks,'' \emph{IEEE Transactions on Evolutionary Computation}, vol.~25, no.~6, pp. 1118--1132, 2021.

\bibitem{sun2019evolving}
Y.~Sun, B.~Xue, M.~Zhang, and G.~G. Yen, ``Evolving deep convolutional neural networks for image classification,'' \emph{IEEE Transactions on Evolutionary Computation}, vol.~24, no.~2, pp. 394--407, 2019.

\bibitem{yang2021gradient}
S.~Yang, Y.~Tian, C.~He, X.~Zhang, K.~C. Tan, and Y.~Jin, ``A gradient-guided evolutionary approach to training deep neural networks,'' \emph{IEEE Transactions on Neural Networks and Learning Systems}, 2021.

\bibitem{taylor2005stirring}
W.~R. Taylor, ``Stirring the primordial soup,'' \emph{Nature}, vol. 434, no. 7034, pp. 705--705, 2005.

\bibitem{miller1953production}
S.~L. Miller, ``A production of amino acids under possible primitive earth conditions,'' \emph{Science}, vol. 117, no. 3046, pp. 528--529, 1953.

\bibitem{kasting1993earth}
J.~F. Kasting, ``Earth's early atmosphere,'' \emph{Science}, vol. 259, no. 5097, pp. 920--926, 1993.

\bibitem{real2017large}
E.~Real, S.~Moore, A.~Selle, S.~Saxena, Y.~L. Suematsu, J.~Tan, Q.~V. Le, and A.~Kurakin, ``Large-scale evolution of image classifiers,'' in \emph{International Conference on Machine Learning}.\hskip 1em plus 0.5em minus 0.4em\relax PMLR, 2017, pp. 2902--2911.

\bibitem{lu2020multiobjective}
Z.~Lu, I.~Whalen, Y.~Dhebar, K.~Deb, E.~D. Goodman, W.~Banzhaf, and V.~N. Boddeti, ``Multiobjective evolutionary design of deep convolutional neural networks for image classification,'' \emph{IEEE Transactions on Evolutionary Computation}, vol.~25, no.~2, pp. 277--291, 2020.

\bibitem{Xiao2017-hq}
H.~Xiao, K.~Rasul, and R.~Vollgraf, ``{Fashion-MNIST}: a novel image dataset for benchmarking machine learning algorithms,'' Aug. 2017.

\bibitem{Krizhevsky2009-st}
A.~Krizhevsky, G.~Hinton, and {Others}, ``Learning multiple layers of features from tiny images,'' 2009.

\end{thebibliography}

% \newpage
% \section{Appendix}



\end{document}